\documentclass[10pt,twocolumn,letterpaper]{article}

\usepackage{cvpr}              

\usepackage[dvipsnames]{xcolor}

\usepackage{float}
\usepackage{amsmath}
\usepackage{color}
\usepackage{tikz}

\def\ie{\emph{i.e.}}
\def\eg{\emph{e.g.}}

\definecolor{cvprblue}{rgb}{0.21,0.49,0.74}
\usepackage[pagebackref,breaklinks,colorlinks,citecolor=cvprblue]{hyperref}

\def\paperID{1112} 
\def\confName{CVPR}
\def\confYear{2024}

\title{Text2Street: Controllable Text-to-image Generation for Street Views}

\author{Jinming Su$^*$, Songen Gu$^*$, Yiting Duan, Xingyue Chen and Junfeng Luo \\
Meituan \\
{\tt\small sujinming0125@gmail.com}
}

\begin{document}
\twocolumn[{%
\maketitle 
\begin{figure}[H] 
\hsize=\textwidth 
\centering \includegraphics[width=1\textwidth]{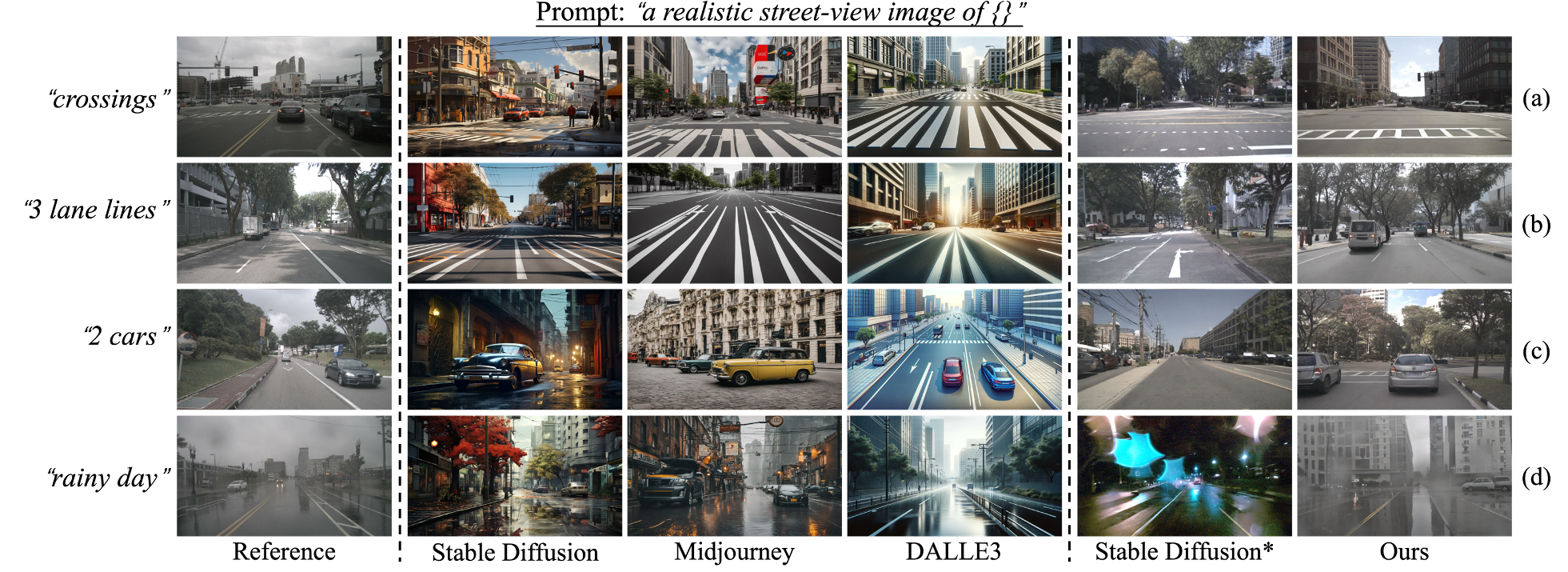}  
\caption{Challenges of text-to-image generation for street views. There are three primary challenges: (1) complex road topology, including road structure in the first row and topological marks in the second row, (2) diverse traffic status, \eg, varying traffic objects in the third row, and (3) various weather conditions like the rainy day in the last row. Note that Reference are original images from nuScenes~\cite{caesar2020nuscenes}, Stable Diffusion~\cite{rombach2022high}/Midjourney~\cite{midjourney}/DALLE3~\cite{dalle3} are tested on their official APIs, and Stable Diffusion$^*$ and Ours are finetuned on nuScenes.} 
\label{fig:motivation}
\end{figure} 
}] 
\let\thefootnote\relax\footnotetext{$^*$ Equal contribution.}



\begin{abstract}
Text-to-image generation has made remarkable progress with the emergence of diffusion models.
However, it is still a difficult task to generate images for street views based on text, mainly because the road topology of street scenes is complex, the traffic status is diverse and the weather condition is various, which makes conventional text-to-image models difficult to deal with. 
To address these challenges, we propose a novel controllable text-to-image framework, named \textbf{Text2Street}. 
In the framework, we first introduce the lane-aware road topology generator, which achieves text-to-map generation with the accurate road structure and lane lines armed with the counting adapter, realizing the controllable road topology generation. 
Then, the position-based object layout generator is proposed to obtain text-to-layout generation through an object-level bounding box diffusion strategy, realizing the controllable traffic object layout generation.
Finally, the multiple control image generator is designed to integrate the road topology, object layout and weather description to realize controllable street-view image generation.
Extensive experiments show that the proposed approach achieves controllable street-view text-to-image generation and validates the effectiveness of the Text2Street framework for street views.
\end{abstract}

\section{Introduction}
\begin{figure*}[t]
\centering
\includegraphics[width=1\textwidth]{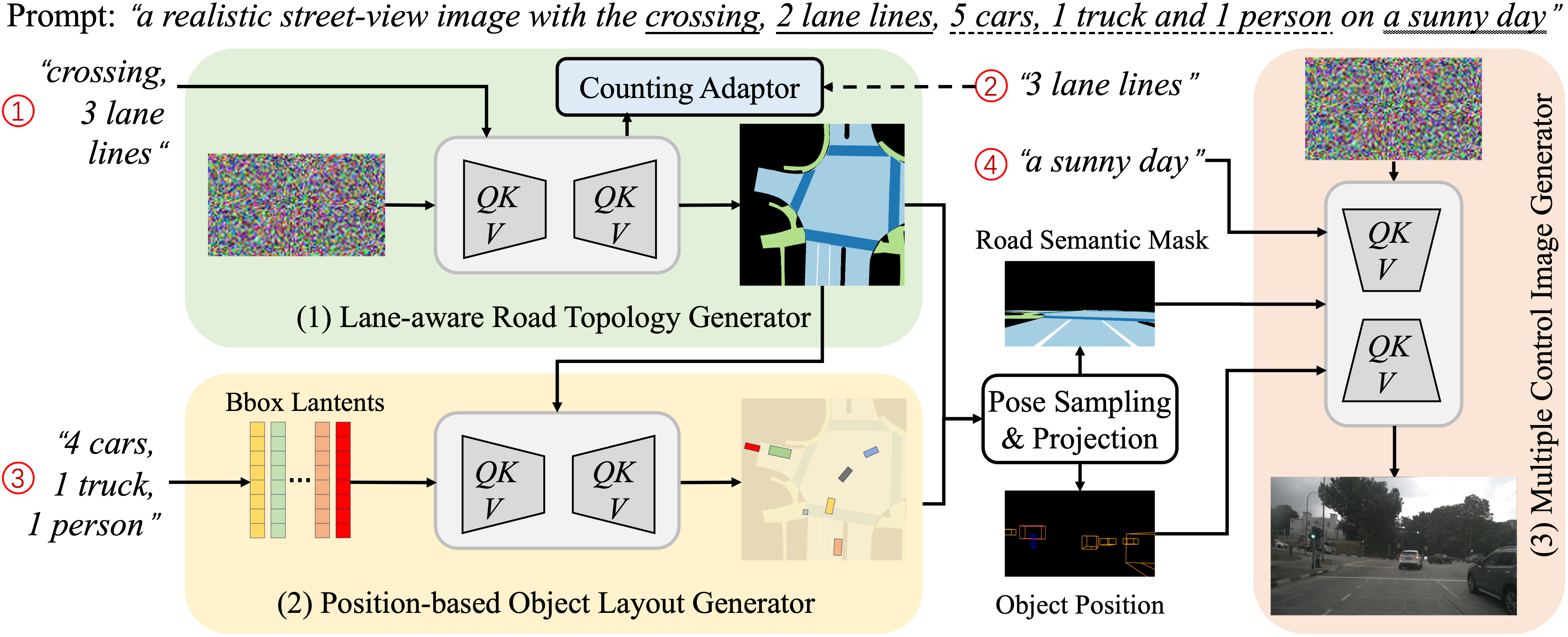} 
\caption{Framework of \textbf{Text2Street}. We begin by introducing the lane-aware road topology generator, which utilizes textual input to create a local semantic map representing the intricate road topology with lane information. Next, we present the position-based object layout generator, which captures the diversity of traffic status and generate the traffic object layout. Subsequently, the road topology and object layout are projected into the camera's perspective through pose sampling. Finally, the projected road topology, object layout, and textual weather description are integrated through the multiple control image generator to produce the ultimate street-view image.}
\label{fig:framework}
\end{figure*} 

Text-to-image generation~\cite{mansimov2015generating,gregor2015draw,reed2016learning}, as an essential task of computer vision that aims to coherent images solely based on textual descriptions. 
In recent years, great efforts~\cite{ramesh2021zero,ramesh2022hierarchical} have been dedicated to text-to-image generation for common scenarios, such as people and objects. Remarkable progress has been achieved, especially with the advent of diffusion models~\cite{ho2020denoising,rombach2022high}. However, it is equally valuable to generate images in specialized domains, including autonomous driving~\cite{marti2019review}, medical image analysis~\cite{chan2020deep}, robot perception~\cite{shi2017design}, among others. Text-to-image generation for street views holds particular importance for data generation in the context of autonomous driving perception and map construction, yet it remains relatively unexplored.

Street-view text-to-image generation, as an underdeveloped task, faces several serious challenges, which can be categorized into three main aspects.
Firstly, generating road topologies that adhere to traffic regulations presents a challenge. On one hand, as depicted in Fig.\ref{fig:motivation} (a), learning the road structure from text-image pairs is hindered by incomplete road structure information in the image, arising from limited imaging angles and frequent occlusions. This complexity makes it challenging for Stable Diffusion$^*$\cite{rombach2022high}, fine-tuned on nuScenes dataset~\cite{caesar2020nuscenes}, to generate expected images. On the other hand, as illustrated in Fig.~\ref{fig:motivation} (b), generating lane lines that both comply with traffic regulations and match the count specified in the text poses a formidable challenge.
Secondly, the representation of traffic status, a crucial element in street-view images, is often achieved through the number of traffic objects present. However, generating a specified number of traffic objects while adhering to motion rules using current models frequently falls short of expectations. As demonstrated in Fig.~\ref{fig:motivation} (c), existing methods tend to lack sensitivity to precise numerical requirements. For instance, while our goal is to generate a road scene with two cars, the actual output from Stable Diffusion$^*$ often includes a significantly higher number of cars.
Lastly, weather conditions are typically contingent upon the scene content, and direct image generation based on these conditions often yields vague or suboptimal outcomes, as depicted in Fig.~\ref{fig:motivation} (d).
Due to the presence of these three challenges, street-view text-to-image generation is a demanding task in computer vision.

To address previously mentioned challenges, we propose a novel controllable text-to-image framework for street views termed as \textbf{Text2Street}, illustrated in Fig.~\ref{fig:framework}. Within this framework, we first introduce the lane-aware road topology generator, which utilizes text descriptions to create a local semantic map representing the intricate road topology. This generator also produces lane lines within the semantic map that conform to specified quantities and traffic regulations through a counting adapter.
Subsequently, we introduce the position-based object layout generator to capture the diverse traffic status. By employing an object-level bounding box diffusion strategy, it generates the traffic object layout based on textual descriptions that adhere to specified quantities and traffic rules.
Finally, the road topology and object layout are projected into the camera's imaging perspective through pose sampling. The projected road topology, object layout, and textual weather description are then integrated using the multiple control image generator to produce the final street-view image. Experimental validation confirms the effectiveness of our proposed method in generating street-view images from textual inputs.

The main contributions of this paper are as follows: 1) We propose a novel controllable text-to-image framework for street views, enabling the controls of road topology, traffic status, and weather conditions based solely on text descriptions. 2) We introduce the lane-aware road topology generator that generates specific road structures as well as lane topologies. 3) We propose the position-based object layout generator, capable of generating a specific number of traffic objects that comply with traffic rules. 4) We propose the multiple control image generator that can integrate road topology, traffic status, and weather conditions to achieve multi-condition image generation.

\section{Related Work}
In this section, we review related works in two aspects.

\subsection{Text-to-image Generation}
In recent years, many methods~\cite{mansimov2015generating,reed2016learning,ramesh2021zero,ramesh2022hierarchical,ho2020denoising,rombach2022high,saharia2022photorealistic,dalle3,ding2021cogview,zhou2021lafite} have been dedicated to dealing with the task of general text-to-image generation. For example, 
AlignDRAW~\cite{mansimov2015generating} iteratively draws patches on a canvas, while attending to the relevant words in the description. 
GAWWN~\cite{reed2016learning} synthesizes images given instructions describing what content to draw in which location based on generative adversarial networks~\cite{goodfellow2014generative}.
DALLE~\cite{ramesh2021zero} describes a simple approach for this text-to-image task based on a transformer that autoregressively models the text and image tokens as a single stream of data.
DALLE2~\cite{ramesh2022hierarchical} proposes a two-stage model: a prior that generates a CLIP~\cite{radford2021learning} image embedding given a text caption, and a decoder that generates an image conditioned on the image embedding.
DDPM~\cite{ho2020denoising} presents high quality image synthesis results using diffusion models~\cite{sohl2015deep}, a class of latent variable models inspired by considerations from nonequilibrium thermodynamics.
Stable Diffusion~\cite{rombach2022high} applies diffusion models training in the latent space of pretrained autoencoders, and turns diffusion models into powerful and flexible generators for general conditioning inputs by introducing cross-attention layers into the model architecture.
These methods have garnered remarkable results in general text-to-image generation. However, their effects in street-view text-to-image tasks is not as commendable.

\subsection{Street-view Image Generation}
There has been a recent surge in the study of methods for street-view image generation. For example, 
SDM~\cite{wang2022semantic} processes semantic layout and noisy image differently. It feeds noisy image to the encoder of the U-Net~\cite{ronneberger2015u} structure while the semantic layout to the decoder by multi-layer spatially-adaptive normalization operators.
BEVGen~\cite{swerdlow2023street} synthesizes a set of realistic and spatially consistent surrounding images that match the bird’s-eye view (BEV) layout of a traffic scenario. BEVGen incorporates a novel cross-view transformation with spatial attention design which learns the relationship between cameras and map views to ensure their consistency.
GeoDiffusion~\cite{chen2023integrating} translates various geometric conditions into text prompts and empower pre-trained text-to-image diffusion models for high-quality detection data generation and is able to encode not only the bounding boxes but also extra geometric conditions such as camera views in self-driving scenes. 
BEVControl~\cite{yang2023bevcontrol} proposes a two-stage generative method that can generate accurate foreground and background contents. 
These methods typically require the input of BEV maps, object bounding boxes, or semantic masks to generate images. However, there is almost no research on generating street-view images relying solely on text. In this paper, we primarily focus on resolving the issue of street-view text-to-image generation.

\section{The Proposed Approach}
To address these challenges (\ie, complex road topology, diverse traffic status, various weather conditions) in street-view text-to-image generation, we introduce~\textbf{Text2Street}, a novel controllable framework illustrated in Fig.~\ref{fig:framework}. 
In this section, details of the approach are described as follows.

\subsection{Overview}
Text2Street takes a street-view description prompt (\eg, ``\emph{a street-view image with the crossing, 3 lanes, 4 cars and 1 truck driving on a sunny day}") as input and generates a corresponding street-view image. Prior to the main process, the input prompt is parsed by a large language model (\eg, GPT-4~\cite{openai2023gpt4}) to extract descriptions of road topology, traffic status, and weather conditions, which are then fed into three main components. The first component is the lane-aware road topology generator, which takes the road topology description (``\emph{crossing, 3 lanes}'') as input and produces a local semantic map. The second component is the position-based object layout generator, which takes the traffic object description from the traffic status (``\emph{4 cars and 1 truck}'') as input and generates traffic object layout. The third component is the multiple control image generator, which takes road topology, object layout, and weather condition descriptions (``\emph{a sunny day}'')  as input, and outputs an image that matches the original street-view description prompt.

\subsection{Lane-aware Road Topology Generator}

\begin{figure}[t]
\centering
\includegraphics[width=1\columnwidth]{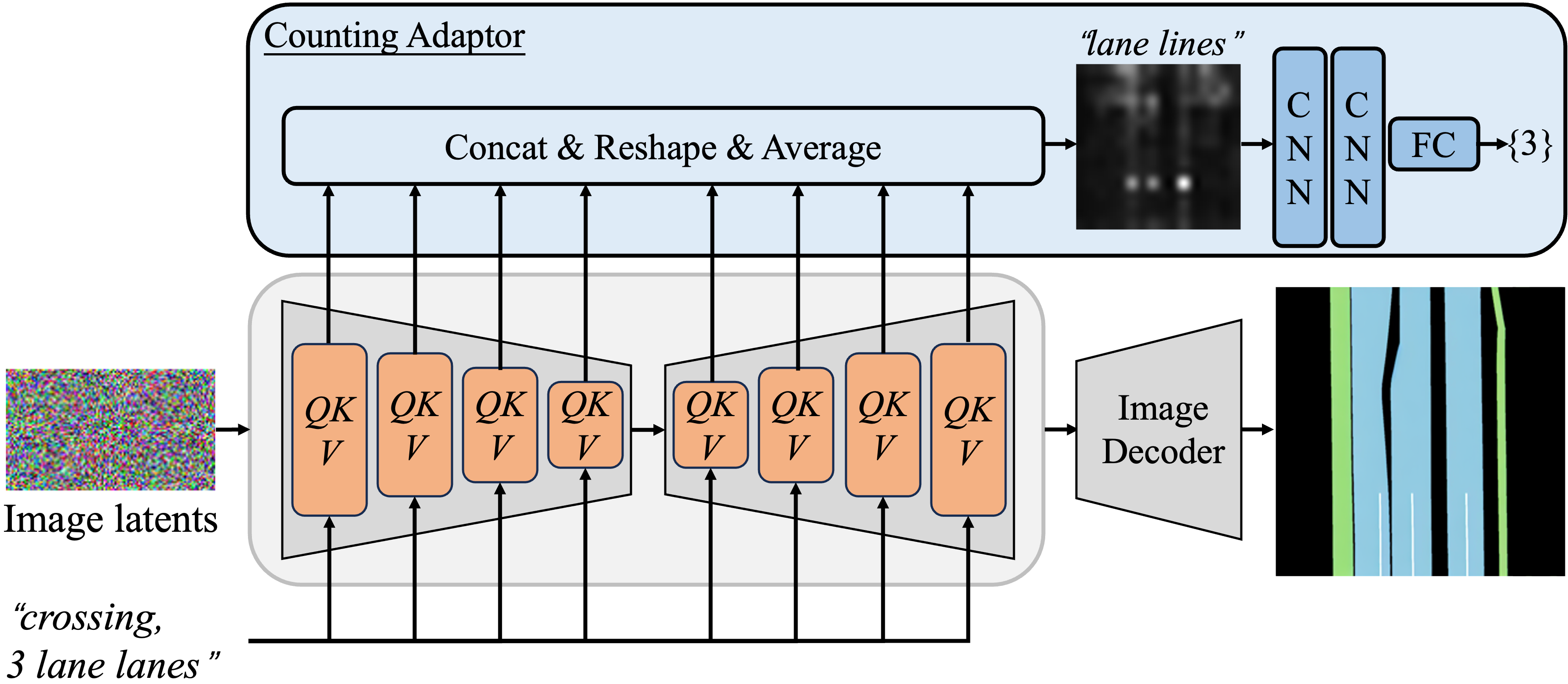}
\vspace{-0.5cm}
\caption{Architecture of the lane-aware road topology generator.}
\label{fig:LRTG}
\end{figure}

For Stable Diffusion~\cite{rombach2022high}, directly generating images that comply with road topology, including road structure and lane topologies, is difficult. To address this, we introduce the lane-aware road topology generator (LRTG), as shown in Fig.~\ref{fig:LRTG}. This generator does not directly produce road images; instead, it first creates a local semantic map describing the road structure, representing a complete regional-level road structure, including drivable areas, intersections, sidewalks, zebra crossings, etc. Simultaneously, to ensure the generated lane lines adhere to traffic regulations (\ie, equidistant and parallel lanes), we characterize and generate lane lines on the semantic map, which is easier and more controllable than generating lane lines directly on perspective-view images. Furthermore, to ensure the number of lane lines aligns with the provided text, we incorporate a counting adapter for the precise generation of a specified number of lane lines. In LRTG, we only generate the semantic map, which serves as a crucial intermediary for street-view images, as further detailed in Section~\ref{chap:MCIG}.

When generating local semantic maps, we utilize Stable Diffusion to encode road topology descriptions based on the CLIP~\cite{radford2021learning} text encoder. Subsequently,  the encoded input is then fed into cross-attention layers of U-Net~\cite{ronneberger2015u} to denoise image latents, ultimately outputting the corresponding semantic map. Consistent with Stable Diffusion, the learning objective is as follows:
\begin{equation}
\begin{aligned}
\mathcal{L}_{SD} = \mathbb{E}_{\mathcal{E}(x),y,\epsilon \sim \mathcal{N}(0,1),t} \left[ \left\| \epsilon - \epsilon_{\theta} (z_t, t, \tau(y)) \right\|^2_{2} \right],
\end{aligned}
\label{eq:loss_sd}
\end{equation}
where $x \in \mathbb{R}^{H \times W \times 3}$ is the given images cropped from labeled semantic maps in RGB space, $\mathcal{E}(\cdot)$ refers to the encoder of pretrained autoencoders~\cite{esser2021taming} and $z=\mathcal{E}(x)$ represents encoded image latents, $z_t$ is from the forward diffusion process at the timestep $t$, $y$ is the text prompt and $\tau(\cdot)$ represents the pretrained CLIP text encoder, the term $\epsilon$ denotes the target noise, and $\epsilon_{\theta}(\cdot)$ signifies the time-conditional U-Net used for predicting the noise.
This manner ensures the reasonable generation of road structures and lane line shapes within the semantic map.

For achieving precise control over the number of lane lines, the counting adapter $f_{CA}$ gathers attention scores from all cross-attention layers of the U-Net. These scores are subsequently reshaped to match the same resolution and then averaged to yield attention features for all tokens. From these attention features, the ones corresponding to tokens ``\emph{lane lines}'' are selected. These selected features $\mathcal{F}_{l}$, undergo further processing through two convolutional layers with the kernel of $3 \times 3$, followed by one fully connected layer, which serves to predict the number of lane lines $N_{l}$. The learning objective for achieving precise control over the number of lane lines is as follows:
\begin{equation}
\begin{aligned}
\mathcal{L}_{CA} = \left\| N_{l} - f_{CA}(\mathcal{F}_{l}) \right\|^2_{2}.
\end{aligned}
\label{eq:loss_ca}
\end{equation}

Based on Eq.~\ref{eq:loss_sd} and~\ref{eq:loss_ca}, LRTG can be jointly optimized to generate the local semantic map, encompassing both road structure and lane lines as required.

\subsection{Position-based Object Layout Generator}

\begin{figure}[t]
\centering
\includegraphics[width=1\columnwidth,width=6.8cm]{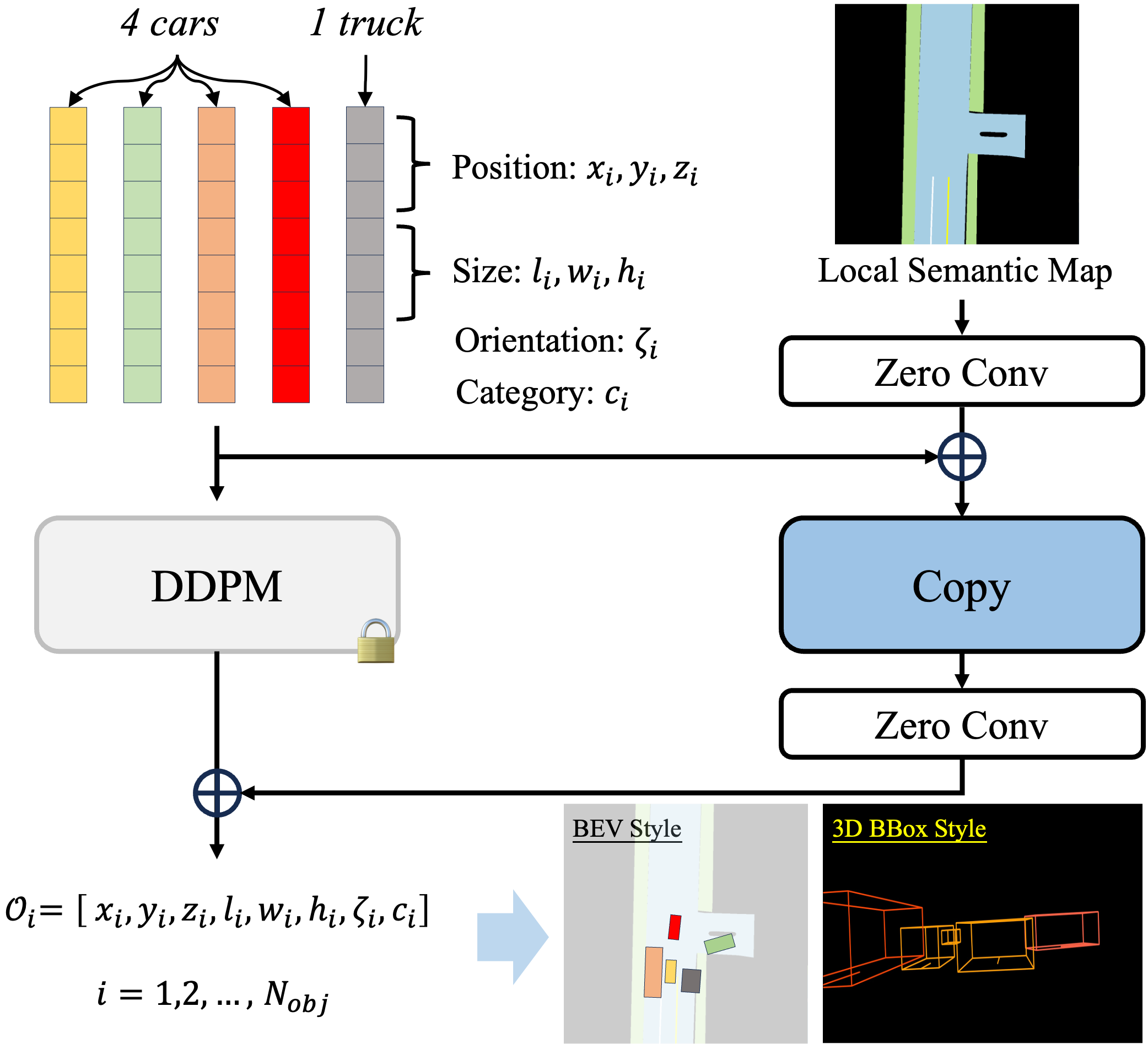}
\caption{Architecture of the position-based object layout generator. Note $\oplus$ means element-wise addition.}
\label{fig:POLG}
\end{figure}

To ensure that generated images can depict diverse traffic conditions, we utilize the large language model to convert traffic status into the number of traffic objects (\eg, car, truck, pedestrian, etc). Then, the position-based object layout generator (POLG) is proposed to create an object layout based on the text description of object quantity, as demonstrated in Fig.~\ref{fig:POLG}. To guarantee a specified number of objects are generated, we incorporate an object-level bounding box diffusion strategy to generate positions of object bounding boxes. Simultaneously, to ensure the generated traffic objects comply with traffic rules, we incorporate the local semantic map from the LRTG into the box diffusion process. With POLG, we generate layout information for traffic objects, which also serves as an intermediary for generating the final street-view images, as introduced in Section~\ref{chap:MCIG}.

In the bounding box diffusion strategy, we first represent traffic objects as position vectors $\mathcal{O}_i = \left[x_i, y_i, z_i, l_i, w_i, h_i, \zeta_i, c_i \right]$ ($i=1, 2, ..., N_{o}$, $N_{o}$ is the number of objects), where $x_i, y_i, z_i$ denote the coordinate of object position, $l_i, w_i, h_i$ represent the object's size with length/width/height, $\zeta_i$ signifies the object's yaw angle, and $c_i$ indicates the object's category. Subsequently, the position vector is diffused based on diffusion models DDPM~\cite{ho2020denoising}. 
Furthermore, to ensure objects adhere to traffic regulations (such as cars must be driven on the road and not against traffic), we use ControlNet~\cite{zhang2023adding}, incorporating the local semantic map from LRTG as a control into the POLG. Ultimately, the learning objective is as follows:
\begin{equation}
\begin{aligned}
\mathcal{L}_{POLG} = \mathbb{E}_{o,m,\epsilon,t} \left[ \left\| \epsilon - \epsilon_{\theta} (o_t, t, \mathcal{C}(m)) \right\|^2_{2} \right],
\end{aligned}
\label{eq:loss_dmc}
\end{equation}
where $o$ represents the position vectors of the objects, $o_t$ is from the forward diffusion process at the timestep $t$, $m$ denotes the local semantic map, and $\mathcal{C}(\cdot)$ signifies the ControlNet. And other symbols are consistent with those in Eq.~\ref{eq:loss_sd}.

Based on Eq.~\ref{eq:loss_dmc}, the layout information of traffic objects that meet the traffic status can be optimized and generated through POLG based on textual descriptions.

\subsection{Multiple Control Image Generator}
\label{chap:MCIG}
\begin{figure}[t]
\centering
\includegraphics[width=1\columnwidth]{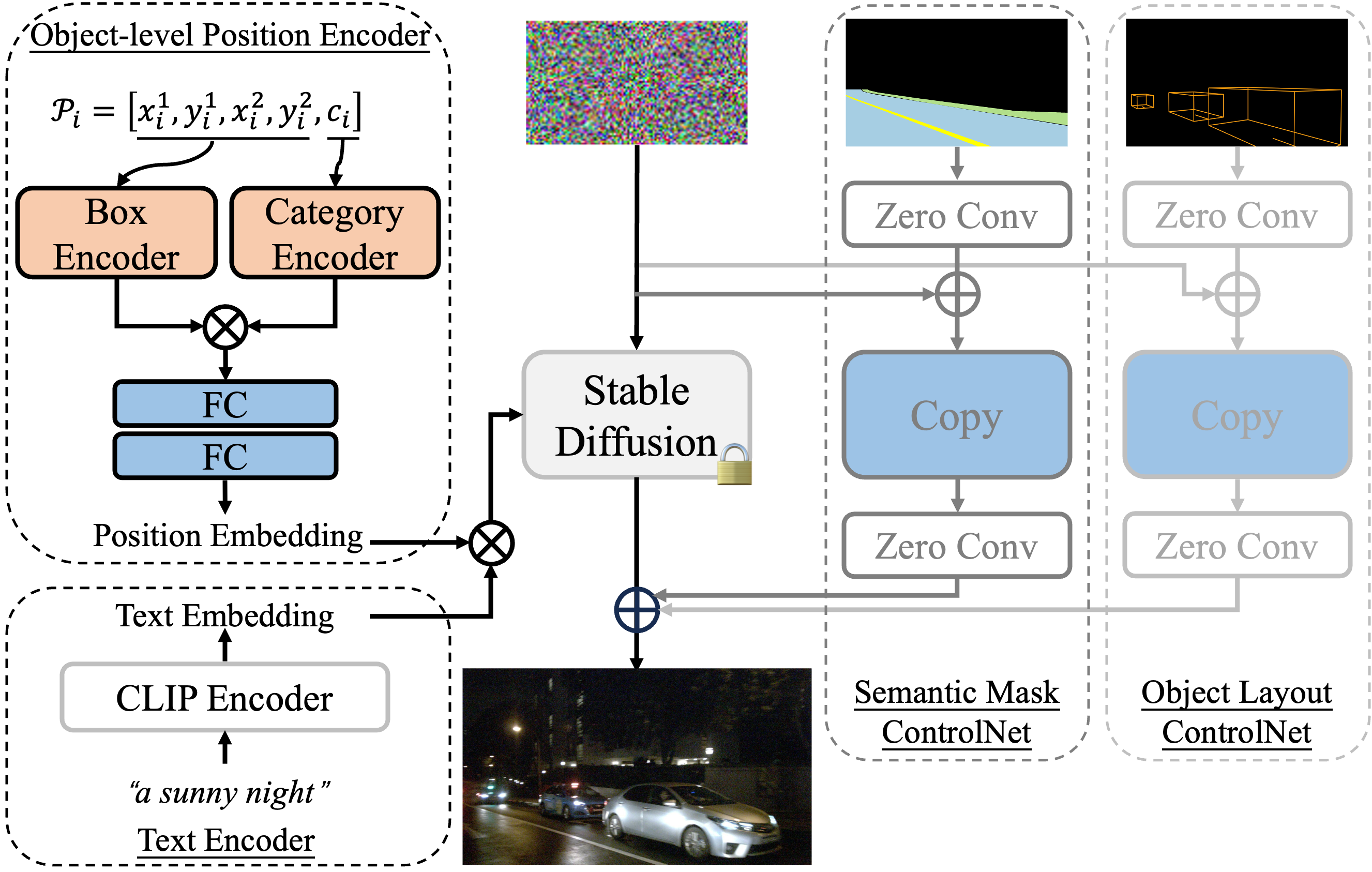}
\vspace{-0.5cm}
\caption{Architecture of the multiple control image generator. Note $\otimes$, $\oplus$ means the concatenation and element-wise addition.}
\label{fig:MCIG}
\end{figure}

To produce images with realistic weather that align with road topology and traffic status, we introduce the multiple control image generator (MCIG), as depicted in Fig.~\ref{fig:MCIG}.

To effectively utilize the previously generated local semantic map and traffic object layout, camera pose sampling and image projection are conducted before these two pieces of information enter MCIG. This results in a 2D road semantic mask $\mathcal{M}_{r}$ and traffic object layout map $\mathcal{M}_{o}$ from a perspective view, as shown in Fig.~\ref{fig:framework}. 
The 2D traffic object layout maps are also represented as 2D traffic object position vectors $\mathcal{P} = \{\mathcal{P}_{i}\}_{i=1}^{N_{o}}=\{\left[x^1_i, y^1_i, x^2_i, y^2_i, c_i \right]\}_{i=1}^{N_{o}}$.
The projection uses a conventional method based on intrinsic and extrinsic transformation, where the intrinsic parameters use fixed camera parameters, and the extrinsic parameters are sampled near the prior camera height.

As depicted in Fig.~\ref{fig:MCIG}, MCIG comprises five modules: object-level position encoder, text encoder, semantic mask ControlNet, object layout ControlNet, and naive Stable Diffusion. The first four modules control image generation based on four different types of information, \ie, 2D traffic position vectors, text describing the weather, 2D road semantic masks, and 2D traffic object layout maps.

The object-level position encoder encodes the 2D traffic object position vectors, including 2D bounding boxes and object categories, represented as:
\begin{equation}
\begin{aligned}
 \mathcal{PE} (\mathcal{P}) = f_{\mathcal{PE}}(\mathcal{BE} \otimes \mathcal{CE}).
\end{aligned}
\label{eq:loss_pe}
\end{equation}
The box encoder maps object bounding boxes to a higher-dimensional space, ensuring that the network can learn higher-frequency mapping functions and focus on the positions of each object. Specifically, the box encoder is an encoding function based on sine and cosine. The mathematical form of the encoding function is as follows:
\begin{equation}
\begin{aligned}
\mathcal{BE}(p) = \left[ \cdots, \text{sin}(2^l \pi p), \text{cos}(2^l \pi p), \cdots \right]^{L-1}_{l=0},
\end{aligned}
\label{eq:loss_be}
\end{equation}
where $\mathcal{BE}(\cdot)$ is applied to each component of the box (\ie, $x^1_i, y^1_i, x^2_i, y^2_i$) of each object $\mathcal{P}_i$, and $L$ is empirically set to 10. 
Simultaneously, the category encoder $\mathcal{CE}$ employs the CLIP text encoder to encode the object category (\eg, ``car''). Subsequently, the box encoding and category encoding are concatenated at the feature embedding dimension of each object. The concatenated features are then mapped to features with the same dimension as the original text encoder's embedding through a two-layer fully connected network $f_{\mathcal{PE}}(\cdot)$, serving as position embeddings.
The text encoder, based on the CLIP text encoder, encodes the weather description text $\mathcal{T}$, resulting in text embeddings. 

The object position encodings and weather text embeddings, upon concatenation at the token dimension, are fed into the cross-attention layer of Stable Diffusion, individually controlling the object position and weather during the image generation.
Simultaneously, the semantic mask ControlNet and object layout ControlNet employ two similar ControlNets, utilizing images (\ie, semantic masks and layout maps) as inputs to control the road topology and object layout during the street-view image generation.
The learning objective function of MCIG is as follows:
\begin{equation}
\begin{aligned}
& \mathcal{L}_{MCIG} = \\
& \mathbb{E}_{\mathbb{P}}\left[ \left\| \epsilon - \epsilon_{\theta} (z_t, t, \mathcal{PE}(\mathcal{P}),\tau(\mathcal{T}),\mathcal{C}(\mathcal{M}_{r}),\mathcal{C}(\mathcal{M}_{o})) \right\|^2_{2} \right], 
\end{aligned}
\label{eq:loss_mcig}
\end{equation}
where  $\mathbb{P}$ is the set of $\{\mathcal{E}(x), \mathcal{P}, \mathcal{T}, \mathcal{M}_{r},\mathcal{M}_{o},\epsilon,t\}$ for convenience of presentation.

Through the optimization of MCIG using Eq.~\ref{eq:loss_mcig}, we obtain street-view images that conform to the initial prompt about road topology, traffic status, and weather conditions.

\section{Experiments and Results}

\subsection{Experimental Setup}

\noindent\textbf{Datasets.}
To validate the performance of the proposed approach, we conduct all experiments on the public autonomous driving dataset nuScenes~\cite{caesar2020nuscenes}. 
nuScenes dataset contains 1,000 street-view scenes (700/150/150 for training/validation/testing, respectively). Each scene comprises approximately 40 frames, with each frame encompassing six RGB images captured by six cameras mounted for panoramic view on the ego vehicle. Additionally, each frame is with a labeled semantic map with 32 semantic categories. 
For the sake of simplicity and clarity, we solely use images captured by the $\mathtt{FRONT}$ camera in all experiments.

\begin{table*}[t]
\centering
\caption{Comparisons with state-of-the-art methods on nuScenes validation dataset. The best result is in {\color{red}{\textbf{bold}}} fonts.}
\setlength{\tabcolsep}{6.1mm}{
\renewcommand\arraystretch{1.1}
\begin{tabular}{c | c c | c c c c}
\hline
& $S_{\text{FID}}$ $\downarrow$ & $S_{\text{CLIP}}$ $\uparrow$ & $S_{\text{road}}$ $\uparrow$ & $S_{\text{lane}}$ $\uparrow$ & $S_{\text{obj}}$ $\uparrow$ & $S_{\text{wea}}$ $\uparrow$\\
\hline
Reference& 0 & 18.00 & 89.67 & 48.30 & 52.33 & 97.00  \\
\hline
Stable Diffusion~\cite{rombach2022high} & 67.91 & 16.38 & 83.50 & 21.33 & 29.83 & 98.50   \\
Stable Diffusion 2.1~\cite{sd2.1} & 63.83 & 16.40 & 83.57 & 29.81 & 31.42 & 98.55  \\ 
Attend-and-Excite~\cite{chefer2023attend} & 69.63 & 16.40 & 83.00 & 30.67 & 31.00  & 98.83 \\
\hline
\textbf{Text2Street (Ours)} & {\color{red}{\textbf{53.92}}}   & {\color{red}{\textbf{17.81}}} & {\color{red}{\textbf{84.17}}} &  {\color{red}{\textbf{35.17}}} & {\color{red}{\textbf{46.33}}} &  {\color{red}{\textbf{99.67}}} \\
\hline
\end{tabular}
}
\label{tab:performance}
\end{table*}

\noindent \textbf{Evaluation Metrics.} 
To comprehensively evaluate the text-to-image generation for street views, we assess the generation results from the image level and attribute level.

For image-level evaluation, we use Fréchet Inception Distance (FID) $S_{\text{FID}}$~\cite{heusel2017gans} to measure image fidelity, and CLIP score $S_{\text{CLIP}}$~\cite{hessel2021clipscore} to image-text alignment. 
Please refer to related works~\cite{heusel2017gans, hessel2021clipscore} for computation details.

In the attribute-level evaluation, we primarily measure the accuracy of text-to-image street-view generation in four aspects: road structure, lane line counting, traffic object counting, and weather conditions. 
For these four metrics, we train four neural networks on nuScenes dataset to evaluate scores of generated images. Specifically, a two-class classifier based on ResNet-50~\cite{he2016deep} is trained for road structure accuracy $S_{\text{road}}$ to distinguish whether the road structure in street-view RGB images is an ``intersection" or ``non-intersection". For the accuracy of lane line counting $S_{\text{lane}}$, a six-class classifier is similarly trained on ResNet-50 to distinguish whether the number of lane lines in street-view RGB images is equal to 0, 1, 2, 3, 4, or $\geq 5$. For the accuracy of traffic object counting $S_{\text{obj}}$, a object detector based on YOLOv5~\cite{Jocher_YOLOv5_by_Ultralytics_2020} is trained to evaluate the number of traffic objects in street-view RGB images. For the accuracy of weather conditions $S_{\text{wea}}$, a four-class classifier is also trained on ResNet-50 to distinguish whether the weather conditions in street-view RGB images are sunny day, sunny night, rainy day, or rainy night. All models are trained on nuScenes training dataset and used as evaluation metrics for attribute-level evaluation of street-view image generation.

\noindent \textbf{Training and Inference.}
During the training phase, we separately train three generators, \ie, lane-aware road topology generator (LRTG), position-based object layout generator (POLG) and multiple control image generator (MCIG). 
LRTG and MCIG are initialized with Stable Diffusion\footnote{https://huggingface.co/runwayml/stable-diffusion-v1-5}~\cite{rombach2022high}, POLG is random initialized based on DDPM\footnote{https://huggingface.co/docs/diffusers/api/pipelines/ddpm}~\cite{ho2020denoising} modified with ControlNet~\cite{zhang2023adding}, and CLIP~\cite{radford2021learning} text encoder are fixed with pretrained weights.
For these three generators, we train them with AdamW~\cite{loshchilov2017decoupled} optimizer for 10 epochs with a learning rate $1e^{-4}$, a batch size of 32. In addition, semantic maps in LRTG are resized to the resolution of $512 \times 512$, and RGB images in MCIG are resized to the resolution of $895 \times 512$. 
In the inference phase, the three generators perform inference sequentially, with the denoising iterations all set to 30 times.

\subsection{Comparisons with state-of-the-art methods}

\begin{figure*}[t]
\centering
\includegraphics[width=1\textwidth,height=14cm]{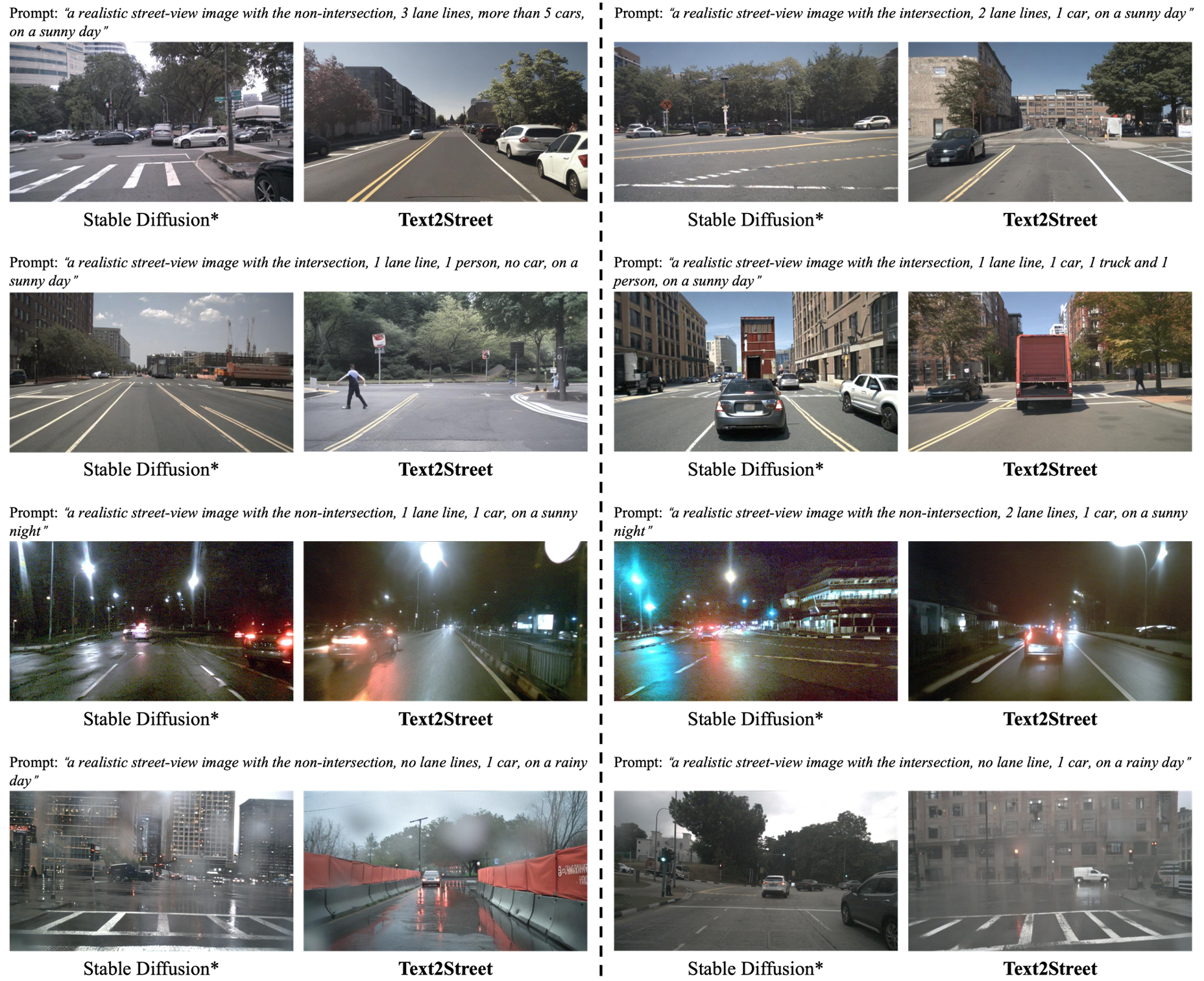}
\vspace{-0.5cm}
\caption{Qualitative comparisons of Stable Diffusion~\cite{rombach2022high} and our approach \textbf{Text2Street}. These two methods are finetuned on nuScenes~\cite{caesar2020nuscenes} dataset. Note that in nuScenes, the double yellow/white lane line counts as one lane line, not two lane lines.}
\label{fig:result}
\end{figure*} 

We compare our approach with several state-of-the-art algorithms in text-to-image generation, including 
Stable Diffusion~\cite{rombach2022high}, Stable Diffusion 2.1~\cite{sd2.1} and Attend-and-Excite~\cite{chefer2023attend} on nuScenes validation dataset, as listed in Tab.~\ref{tab:performance}. These methods are all finetuned on nuScenes training dataset.
Note that we have also listed the performance on the nuScenes validation dataset as the ``Reference''. 

Comparing the proposed method with state-of-the-art methods, we can see that our method consistently outperforms other methods across almost all metrics from Tab.~\ref{tab:performance}.
Significantly, our method outperforms all others on attribute-level metrics (\ie, $S_{\text{road}}$, $S_{\text{lane}}$, $S_{\text{obj}}$ and $S_{\text{wea}}$), demonstrating its superior controllability for fine-grained text-to-image street-view image generation. Specifically, our method shows a obviously 4.50\%, 14.91\% improvement on metric $S_{\text{lane}}$ and $S_{\text{obj}}$ compared to the second best performance. Additionally, our method also performs better on image-level metrics (\ie, $S_{\text{FID}}$ and $S_{\text{CLIP}}$), reflecting its superior overall generation quality and image-text consistency. Overall, these observations validate the effectiveness of our proposed method for controllable image generation for street views.

Visual examples generated by our method are illustrated in Fig.~\ref{fig:result}.
From Fig.~\ref{fig:result}, it is evident that our method yields superior results in dealing with varying road structures (1st and 4th rows), different numbers of lane lines (1st and 3rd rows), diverse numbers of traffic objects (1st and 2nd rows), and various weather conditions (2nd and 3rd rows) compared to other methods. This indicates that our method can effectively generate street-view images only based on text, and also implies its controllability and superiority in street-view text-to-image generation.

\begin{figure*}[t]
\centering
\includegraphics[width=1\textwidth]{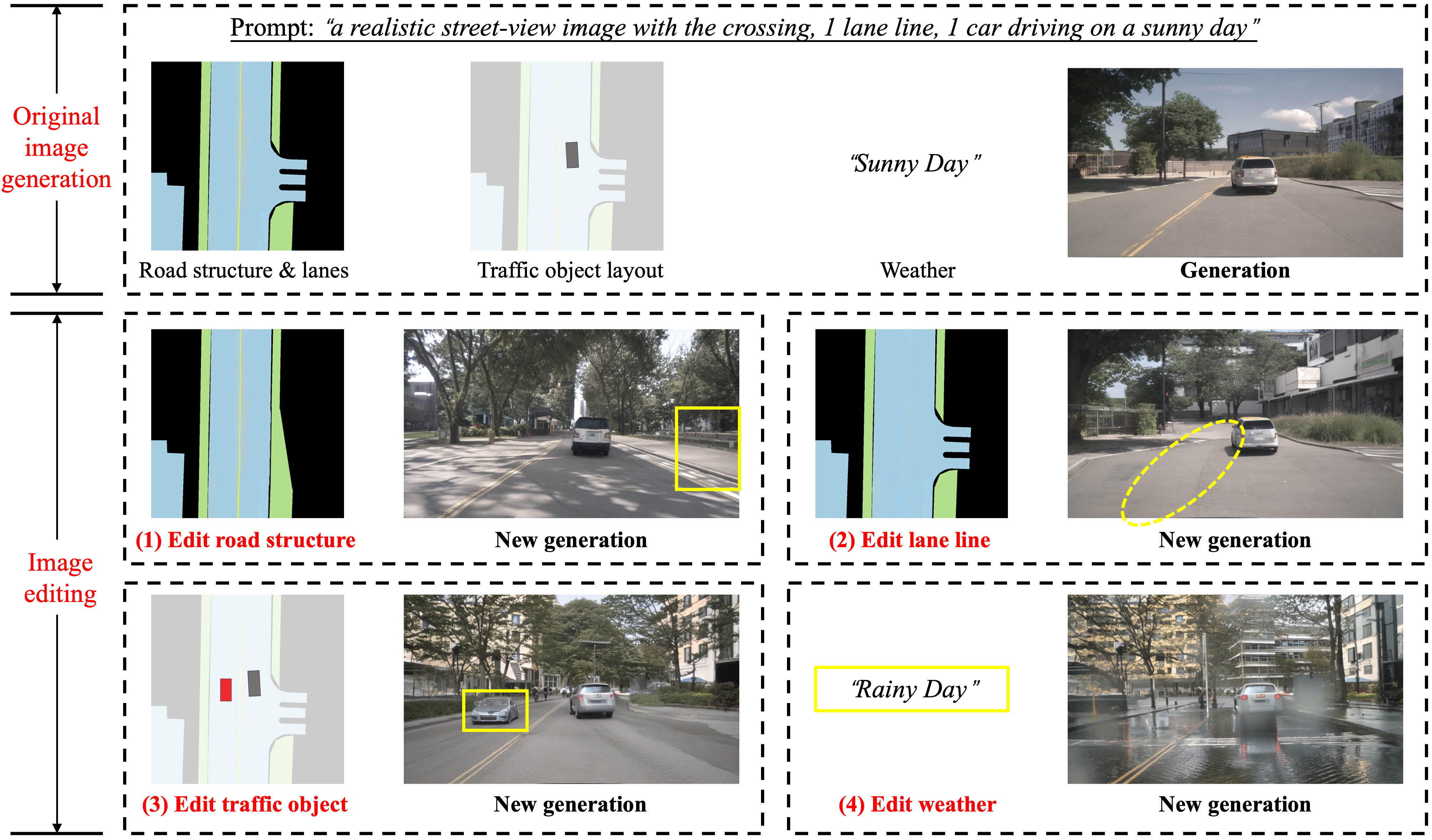}
\vspace{-0.5cm}
\caption{Examples of different image editing operations by our approach.}
\label{fig:editing}
\end{figure*} 

\begin{table*}[t]
\centering
\caption{Performance of different settings of the proposed method on nuScenes validation dataset.}
\setlength{\tabcolsep}{3.5mm}{
\renewcommand\arraystretch{1.05}
\begin{tabular}{c | c c c | c c | c c c c}
\hline
& LRTG & POLG & MCIG & $S_{\text{FID}}$ $\downarrow$ & $S_{\text{CLIP}}$ $\uparrow$ & $S_{\text{road}}$ $\uparrow$ & $S_{\text{lane}}$ $\uparrow$ & $S_{\text{obj}}$ $\uparrow$ & $S_{\text{wea}}$ $\uparrow$\\
\hline
Baseline & & &\checkmark & 67.91 & 16.38 & 83.50 & 21.33 & 29.83 & 98.50   \\
A$_1$ & \checkmark & &\checkmark & 60.45 & {\color{red}{\textbf{17.82}}} & {\color{red}{\textbf{85.17}}} & 33.16 & 30.00 & 98.50  \\
A$_2$ & \checkmark & &\checkmark & 59.85 & {\color{red}{\textbf{17.82}}} & 84.67 & 34.01 & 31.50 & 98.53  \\
B &  & \checkmark  & \checkmark &  59.60& 17.81 & 84.40  & 29.33 & 41.50 & 98.67  \\
\hline
C (\textbf{Text2Street}) & \checkmark & \checkmark & \checkmark & {\color{red}{\textbf{53.92}}}   & 17.81 & 84.17 &  {\color{red}{\textbf{35.17}}} & {\color{red}{\textbf{46.33}}} &  {\color{red}{\textbf{99.67}}} \\
\hline
\end{tabular}
}
\label{tab:performance_ablation}
\end{table*}

\subsection{Ablation Analysis}
To assess the effectiveness of individual components, we carry out ablation experiments on nuScenes validation dataset, comparing the performance variations within the proposed approach.

Fistly, to validate the effectiveness of the lane-aware road topology generator (LRTG), we introduced three models for ablation comparison. The first model, termed as ``Baseline", is a naive multiple control image generator (MCIG) only with only the text encoder, which actually is a Stable Diffusion model. The second model, named ``A$_1$", is based on the ``Baseline" with the addition of LRTG excluding lane line control. The third model, ``A$_2$", adds LRTG with lane line control to the first model. The comparison of these three models is presented in the first three rows of Tab.~\ref{tab:performance_ablation}. It can be observed that the introduction of road structure control (``A$_1$") significantly improves the $S_{\text{road}}$ metric, and the addition of both road structure and lane lines (``A$_2$'') controls further enhances both $S_{\text{road}}$ and $S_{\text{lane}}$ metrics. This confirms the effectiveness of LRTG in controlling road topology.

Secondly, to validate the effectiveness of the position-based object layout generator (POLG), we add POLG to ``Baseline", termed as ``B''. 
Comparing the first and fourth rows of Tab.~\ref{tab:performance_ablation}, it is evident that the inclusion of POLG significantly improves the metric $S_{\text{obj}}$, demonstrating the control capability of POLG in traffic object generation.

Thirdly, to verify the compatibility of different modules, we also list the model ``C'' (\ie, Text2Street), which combines all three modules. As can be seen from the last row of Tab.~\ref{tab:performance_ablation}, ``C" achieves the best performance across all metrics, confirming the compatibility among different modules.

\subsection{Text-to-image Generation for Object Detection}

\begin{table}[t]
\centering
\caption{Performance of YOLOv5 without/with the data augmentation of our method on nuScenes validation dataset.}
\setlength{\tabcolsep}{1 mm}{
\renewcommand\arraystretch{1.1}
\begin{tabular}{c | c c c}
\hline
& mAP $\uparrow$ & Precision $\uparrow$ & Recall $\uparrow$ \\
\hline
YOLOv5 & 46.30 &  81.77 & 70.91 \\
\hline
YOLOv5 with Text2Street & {\color{red}{\textbf{47.83}}}  & {\color{red}{\textbf{84.45}}} & {\color{red}{\textbf{72.37}}} \\
\hline
\end{tabular}
}
\label{tab:object_detection}
\end{table}

To demonstrate the utility of street-view text-to-image generation for downstream tasks, we select object detection as a representative task. We use the proposed Text2Street to generate 30,000 images based on random prompts as a supplement to the original training data to train YOLOv5 on the nuScenes dataset, as listed in Tab.~\ref{tab:object_detection}. The results indicate that the images generated by our method are beneficial for downstream street-view tasks, highlighting the potential of the street-view text-to-image generation.

\subsection{Image Editing}
In addition to street-view text-to-image generation, our approach also allows for modifications to local semantic maps, object layouts, or text, enabling the editing of road structures, lane lines, object layouts, and weather conditions in the originally generated RGB images, as depicted in Fig.~\ref{fig:editing}.

\section{Conclusion}
In this paper, we propose a novel controllable text-to-image generation framework for street views. In this framework, we design the lane-aware road topology generator to exert control over the road topology in a text-to-map manner. Additionally, the position-based object layout generator is proposed to control the layout of traffic objects through a text-to-layout manner. Moreover, the multiple control image generator is built to integrate multiple controls to generate street-view images. Empirical results substantiate the effectiveness of our proposed approach.

{
    \small
    \bibliographystyle{ieeenat_fullname}
    \bibliography{main}
}


\end{document}